\documentclass[10pt,twocolumn,letterpaper]{article}

\usepackage{cvpr}      

\usepackage{tikz}
\usetikzlibrary{arrows.meta, positioning}
\usepackage{graphicx} 
\usepackage{ragged2e}

\definecolor{cvprblue}{rgb}{0.21,0.49,0.74}
\usepackage[pagebackref,breaklinks,colorlinks,allcolors=cvprblue]{hyperref}

\def\paperID{61} 
\def\confName{CVPR}
\def\confYear{2026}

\title{NTIRE 2026 The 3rd Restore Any Image Model (RAIM) Challenge: AI Flash Portrait (Track 3)}

\author{%
\let\and\\%
Ya-nan Guan \hspace{1.7em} Shaonan Zhang \hspace{1.7em} Hang Guo \hspace{1.7em} Yawen Wang \hspace{1.7em} Xinying Fan \and
Tianqu Zhuang \hspace{1.7em} Jie Liang \hspace{1.7em} Hui Zeng \hspace{1.7em} Guanyi Qin \hspace{1.7em} Lishen Qu \and
Tao Dai \hspace{1.7em} Shu-Tao Xia \hspace{1.7em} Lei Zhang \hspace{1.7em} Radu Timofte \hspace{1.7em} Bin Chen \and
Yuanbo Zhou \hspace{1.7em} Hongwei Wang \hspace{1.7em} Qinquan Gao \hspace{1.7em} Tong Tong \hspace{1.7em} Yanxin Qian \and
Lizhao You \hspace{1.7em} Jingru Cong \hspace{1.7em} Lei Xiong \hspace{1.7em} Shuyuan Zhu \hspace{1.7em} Zhi-Qiang Zhong \and
Kan Lv \hspace{1.7em} Yang Yang \hspace{1.7em} Kailing Tang \hspace{1.7em} Minjian Zhang \hspace{1.7em} Zhipei Lei \and
Zhe Xu \hspace{1.7em} Liwen Zhang \hspace{1.7em} Dingyong Gou \hspace{1.7em} Yanlin Wu \hspace{1.7em} Cong Li \and
Xiaohui Cui \hspace{1.7em} Jiajia Liu \hspace{1.7em} Guoyi Xu \hspace{1.7em} Yaoxin Jiang \hspace{1.7em} Yaokun Shi \and
Jiachen Tu \hspace{1.7em} Liqing Wang \hspace{1.7em} Shihang Li \hspace{1.7em} Bo Zhang \hspace{1.7em} Biao Wang \and
Haiming Xu \hspace{1.7em} Xiang Long \hspace{1.7em} Xurui Liao \hspace{1.7em} Yanqiao Zhai \hspace{1.7em} Haozhe Li \and
Shijun Shi \hspace{1.7em} Jiangning Zhang \hspace{1.7em} Yong Liu \hspace{1.7em} Kai Hu \hspace{1.7em} Jing Xu \and
Xianfang Zeng \hspace{1.7em} Yuyang Liu \hspace{1.7em} Minchen Wei
}

\begin{document}
\maketitle

\begin{abstract}
In this paper, we present a comprehensive overview of the NTIRE 2026 3rd Restore Any Image Model (RAIM) challenge, with a specific focus on Track 3: AI Flash Portrait. Despite significant advancements in deep learning for image restoration, existing models still encounter substantial challenges in real-world low-light portrait scenarios. Specifically, they struggle to achieve an optimal balance among noise suppression, detail preservation, and faithful illumination and color reproduction. To bridge this gap, this challenge aims to establish a novel benchmark for real-world low-light portrait restoration. We comprehensively evaluate the proposed algorithms utilizing a hybrid evaluation system that integrates objective quantitative metrics with rigorous subjective assessment protocols. For this competition, we provide a dataset containing 800 groups of real-captured low-light portrait data. Each group consists of a 1K-resolution low-light input image, a 1K ground truth (GT), and a 1K person mask. This challenge has garnered widespread attention from both academia and industry, attracting over 100 participating teams and receiving more than 3,000 valid submissions. This report details the motivation behind the challenge, the dataset construction process, the evaluation metrics, and the various phases of the competition. The released dataset and baseline code for this track are publicly available from the same \href{https://github.com/zsn1434/AI_Flash-BaseLine/tree/main}{GitHub repository}, and the official challenge webpage is hosted on \href{https://www.codabench.org/competitions/12885/}{CodaBench}.
\end{abstract}    
\section{Introduction}
\label{sec:intro}

In the realm of mobile computational photography, capturing high-quality portraits in low-light environments remains a formidable challenge. Constrained by limited sensor sizes and insufficient light intake, low-light portraits are typically plagued by severe noise, color distortion, and significant loss of fine details. To address this issue, the task of AI Flash Portrait has emerged. This novel task aims to map low-light portraits captured with weak flash effects to visually stunning portraits exhibiting professional, strong flash illumination and superior aesthetic quality. Accomplishing this requires algorithms to go beyond mere physical-level illumination enhancement and denoising; they must also operate at an aesthetic level to preserve high-frequency details, faithfully reproduce natural skin tones, and maintain background visual comfort alongside overall scene balance.

However, existing low-level vision and image generation paradigms exhibit pronounced limitations when tackling this highly composite task.

\noindent\textbf{Limitations of Traditional Low-Light Image Enhancement (LLIE)}. The majority of LLIE methods focus primarily on global luminance elevation and are typically trained on synthetic or globally brightened datasets \cite{Xiong2017MSRCR, Wei2018DeepRD, Lv2018MBLLEN, Zhang2019KindlingDarkness}. When applied to human portraits, these approaches often induce skin tone distortion, flatten facial lighting, and amplify severe background noise. Fundamentally, they lack a deep understanding of portrait aesthetics and the physical attenuation characteristics of flash lighting.

\noindent\textbf{Challenges in Real-World Image Restoration}. Although recent studies have attempted to model real-world noise and degradation, the degradation process in authentic scenarios is exceedingly complex \cite{wang2021realesrgan, abdelhamed2018sidd}. Purely synthetic data fails to simulate the intricate, non-linear illumination shifts involved in transitioning from a weak flash to a strong flash. Consequently, the absence of high-quality, real-world paired data-specifically retouched by professional designers-severely impedes the deployment of such models in practical industrial applications.

\noindent\textbf{Shortcomings of Face Restoration and Portrait Generation}.Current state-of-the-art face restoration models (e.g., GAN or Diffusion-based methods) excel at recovering facial details but are generally confined to localized processing. In low-light scenarios, producing an exceptionally sharp and bright face against a noisy or poorly illuminated background inevitably results in a severe cut-and-paste artifact (i.e., spatial disharmony) and breaks the scene balance \cite{wang2021gfpgan, zhou2022codeformer}.

\noindent\textbf{The Dilemma of Image Quality Assessment (IQA) in Generative Tasks}. Traditional objective metrics (e.g., PSNR, SSIM), and even perceptual metrics (e.g., LPIPS), are inadequate at fully capturing human perception of aesthetics and naturalness \cite{zhang2018lpips, wang2004ssim, gu2020pipal}. Particularly when evaluated against high-quality, designer-retouched reference images, models that produce overly smoothed or excessively sharpened outputs may deceptively achieve high objective scores while yielding remarkably poor visual fidelity.

To bridge the substantial gap between academic research and industrial application in low-light portrait computational photography, the Y-Lab of The OPPO Research Institute, the College of Computer Science and Software Engineering at Shenzhen University, the Visual Computing Lab (VC-Lab) at The Hong Kong Polytechnic University, and Nankai University jointly organized the 3rd Restore Any Image Model (RAIM) Challenge: AI Flash Portrait, in conjunction with the NTIRE 2026 workshop. The primary objectives of this challenge are as follows:

\begin{enumerate}[label=\arabic*.]
    \item To establish a novel benchmark for real-world low-light portrait restoration and aesthetic enhancement by providing high-quality, real-world paired data meticulously retouched by professional designers.
    \item To formulate a comprehensive evaluation protocol that seamlessly integrates region-aware objective metrics with rigorous expert blind-testing for subjective assessment.
    \item To encourage the development of robust solutions that simultaneously achieve exceptional portrait rendering and harmonious overall scene quality, thereby facilitating the deployment of advanced algorithms in practical, real-world applications.
\end{enumerate}

This challenge is one of the challenges associated with the NTIRE 2026 Workshop~\footnote{\url{https://www.cvlai.net/ntire/2026/}} on:
Deepfake detection~\cite{ntire26deepfake}, 
high-resolution depth~\cite{ntire26hrdepth},
multi-exposure image fusion~\cite{ntire26raim_fusion}, 
AI flash portrait~\cite{ntire26raim_portrait}, 
professional image quality assessment~\cite{ntire26raim_piqa},
light field super-resolution~\cite{ntire26lightsr},
3D content super-resolution~\cite{ntire263dsr},
bitstream-corrupted video restoration~\cite{ntire26videores},
X-AIGC quality assessment~\cite{ntire26XAIGCqa},
shadow removal~\cite{ntire26shadow},
ambient lighting normalization~\cite{ntire26lightnorm},
controllable Bokeh rendering~\cite{ntire26bokeh},
rip current detection and segmentation~\cite{ntire26ripdetseg},
low light image enhancement~\cite{ntire26llie},
high FPS video frame interpolation~\cite{ntire26highfps},
Night-time dehazing~\cite{ntire26nthaze,ntire26nthaze_rep},
learned ISP with unpaired data~\cite{ntire26isp},
short-form UGC video restoration~\cite{ntire26ugcvideo},
raindrop removal for dual-focused images~\cite{ntire26dual_focus},
image super-resolution (x4)~\cite{ntire26srx4},
photography retouching transfer~\cite{ntire26retouching},
mobile real-word super-resolution~\cite{ntire26rwsr},
remote sensing infrared super-resolution~\cite{ntire26rsirsr},
AI-Generated image detection~\cite{ntire26aigendet},
cross-domain few-shot object detection~\cite{ntire26cdfsod},
financial receipt restoration and reasoning~\cite{ntire26finrec},
real-world face restoration~\cite{ntire26faceres},
reflection removal~\cite{ntire26reflection},
anomaly detection of face enhancement~\cite{ntire26anomalydet},
video saliency prediction~\cite{ntire26videosal},
efficient super-resolution~\cite{ntire26effsr},
3d restoration and reconstruction in adverse conditions~\cite{ntire26realx3d},
image denoising~\cite{ntire26denoising},
blind computational aberration correction~\cite{ntire26aberration},
event-based image deblurring~\cite{ntire26eventblurr},
efficient burst HDR and restoration~\cite{ntire26bursthdr},
low-light enhancement: `twilight Cowboy'~\cite{ntire26twilight},
and efficient low light image enhancement~\cite{ntire26effllie}.

\renewcommand{\thefootnote}{}
\footnotetext{Ya-nan Guan, Shaonan Zhang, Hang Guo, Yawen Wang, Xinying Fan, Tianqu Zhuang, Jie Liang, Hui Zeng, Guanyi Qin, Lishen Qu, Tao Dai, Shu-Tao Xia, Lei Zhang and Radu Timofte are the organizers of the NTIRE 2026 challenge, and other authors are the participants.}
\footnotetext{The Appendix lists the authors’ teams and affiliations.}

\section{NTIRE 2026 the 3rd RAIM Challenge}
\label{sec:challenge}

\subsection{Training Data}

During Phase 1, the organizers released 600 groups of paired data with ground truth to facilitate model development by the participants. Each data group strictly comprises three components: a 1K-resolution low-light input image, a corresponding 1K-resolution GT image, and a 1K-resolution person mask.

It is crucial to emphasize that the input images in our dataset are not captured in extreme, pitch-black environments; rather, they are real-world photographs captured with a weak flash effect. Conversely, the corresponding GT images are high-quality reference photographs meticulously retouched by professional visual designers to exhibit the aesthetic appeal of a strong, studio-level flash. Consequently, this challenge fundamentally transcends traditional low-light enhancement, representing a synergistic combination of physical illumination enhancement and subjective aesthetic rendering.

\begin{figure}[t]
    \centering
    \includegraphics[width=\columnwidth]{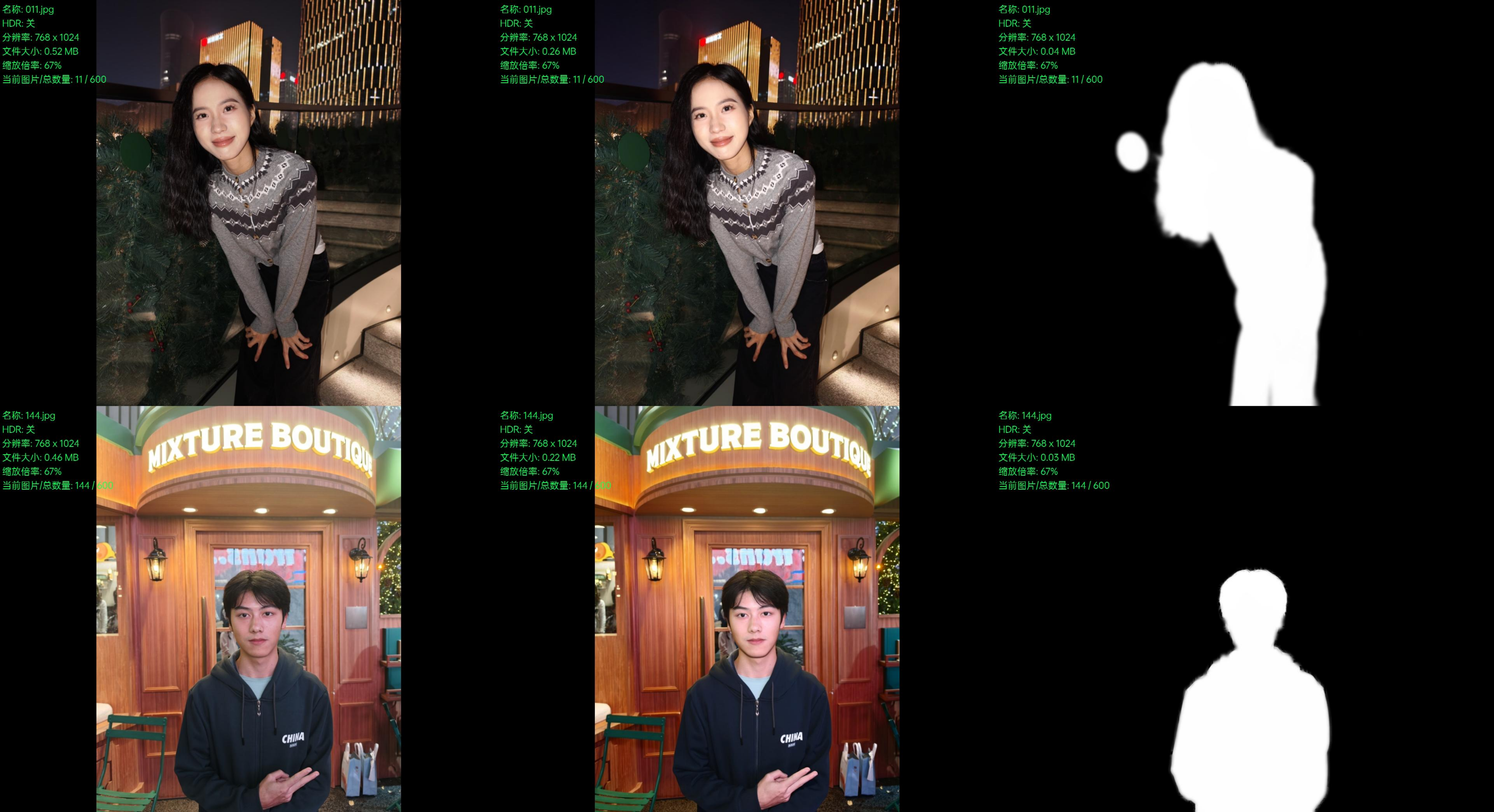}
    \caption{Two representative data pairs from the 600 groups of Phase 1 training data. Each pair comprises a 1K-resolution low-light input image, a corresponding 1K-resolution flash portrait ground truth (GT), and a 1K-resolution person mask.}
    \label{fig:01}
\end{figure}

The training data is publicly accessible via the links provided by the organizers. Furthermore, the challenge permits participants to utilize any publicly available external datasets and pre-trained models for training, without being restricted solely to the provided data. However, the use of any such external resources must be explicitly detailed and disclosed in their final fact sheets and reports. Two representative data pairs are illustrated in Figure \ref{fig:01}.

\subsection{Validation and Test Data}

\subsubsection{Validation Data for Phase 2}

\begin{figure}[t]
    \centering
    \includegraphics[width=\columnwidth]{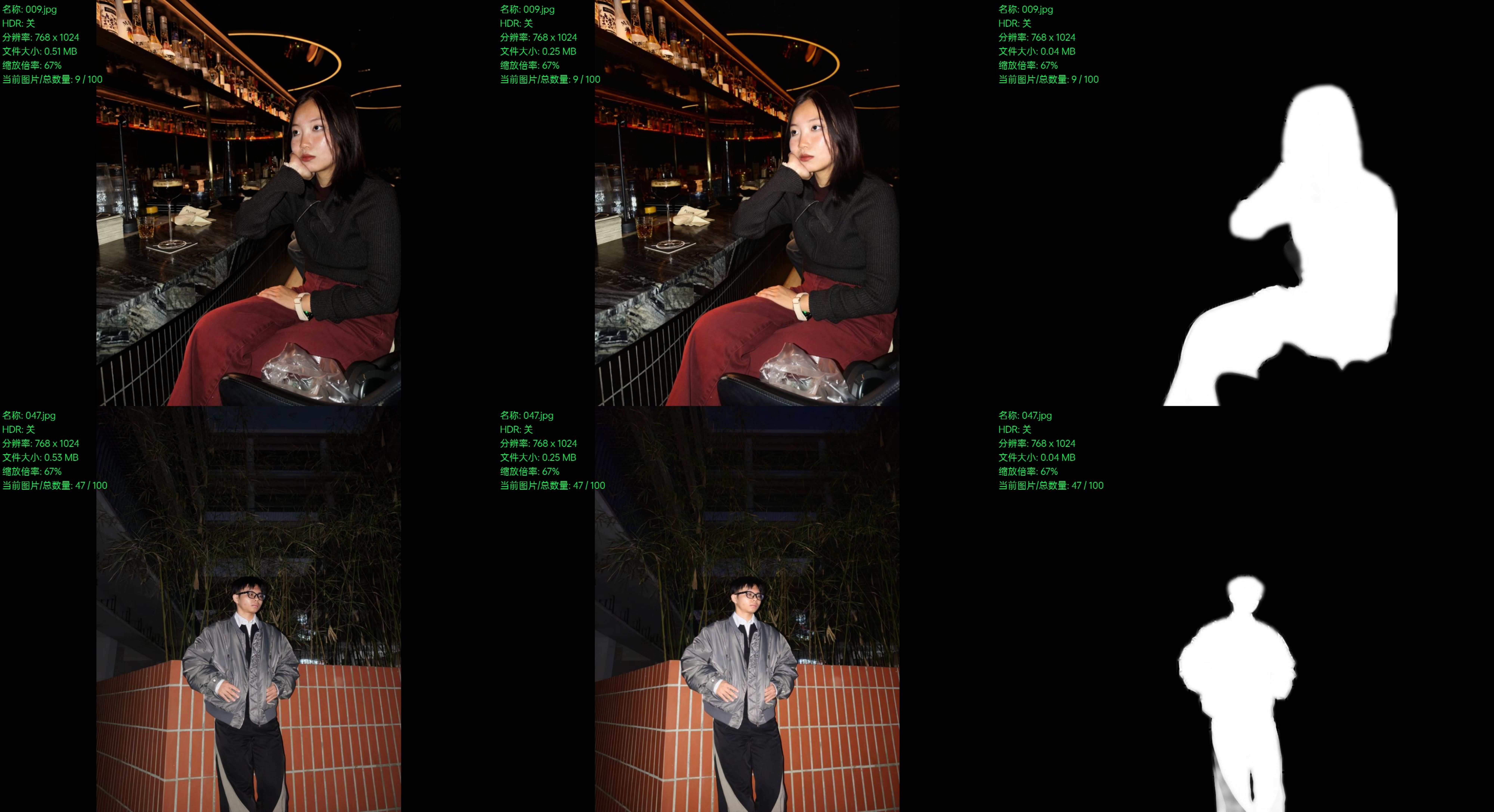}
    \caption{Two representative data pairs from the 100 groups of Phase 2 validation data. Similar to Phase 1, each group includes a 1K low-light input image, a GT, and a person mask. Compared to Phase 1, this split introduces more challenging scenarios, such as distant portraits, to improve the discriminative ability of the evaluation.}
    \label{fig:02}
\end{figure}

During the online validation phase, we released 100 groups of test data, each comprising a 1K-resolution low-light input image and its corresponding 1K person mask. To rigorously evaluate the models' generalization capabilities against unseen, real-world degradations, the high-quality reference ground truths for this split were deliberately withheld. Participants utilized this dataset to conduct inference and subsequently submitted their restored outputs to the online CodaBench server to obtain immediate scoring feedback derived from objective metrics. This online evaluation mechanism empowered participants to continuously monitor their models' performance and iteratively refine their network architectures and hyperparameters, even in the absence of paired GT data. Two representative validation samples are illustrated in Figure \ref{fig:02}.

\subsubsection{Test Data for Phase 3}

In practical industrial deployment scenarios, models are frequently required to process extreme and entirely unseen degradation distributions. Driven by this motivation, for the final expert evaluation phase, we meticulously constructed an additional hidden test set comprising 100 sample groups, which maintains a degradation difficulty distribution consistent with that of Phase 2. To strictly guarantee the absolute fairness and full reproducibility of the final assessment, this test set is kept entirely confidential during the challenge.

\begin{figure}[t]
    \centering
    \includegraphics[width=\columnwidth]{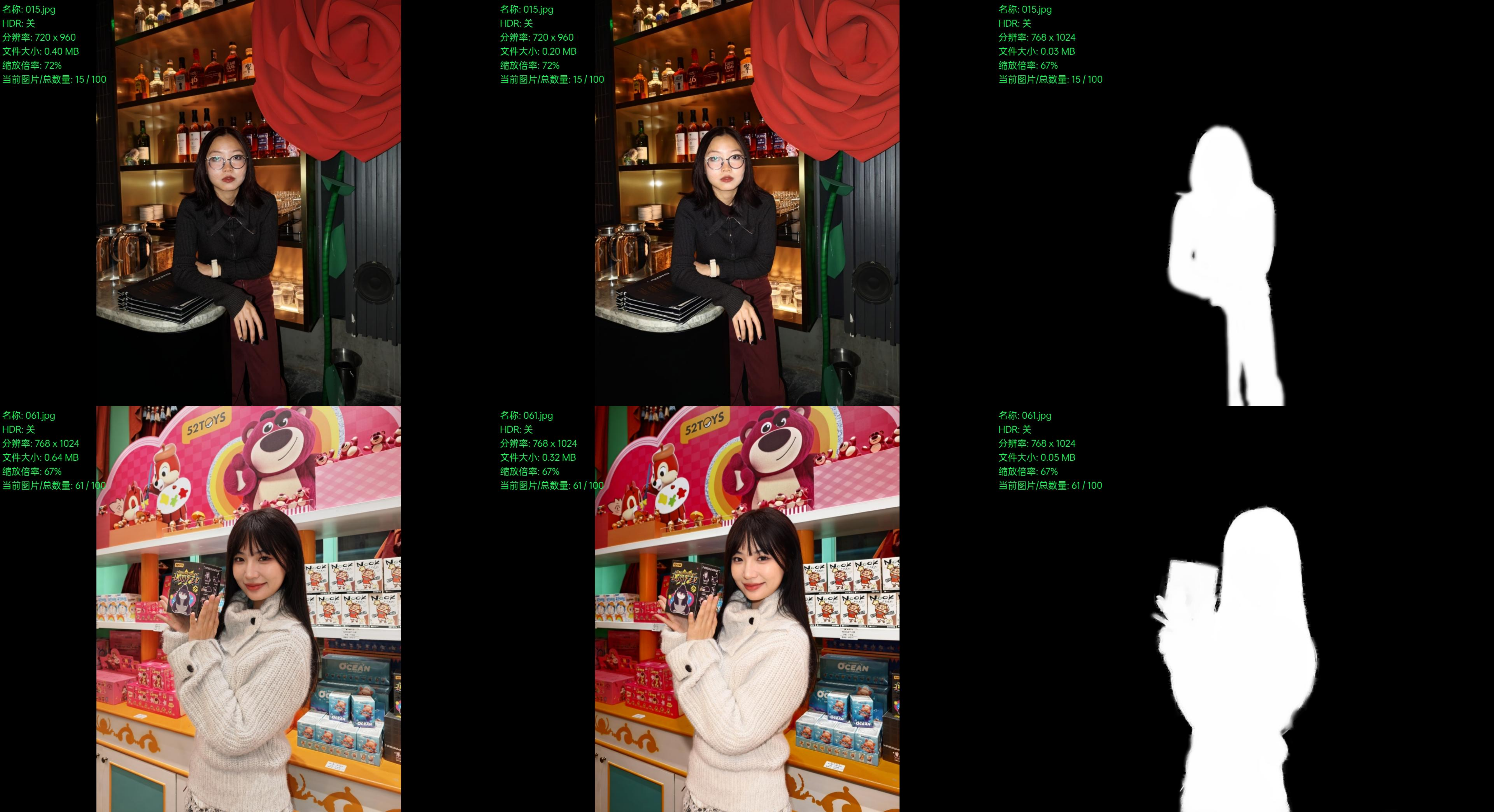}
    \caption{Two representative data pairs from the 100 groups of Phase 3 hidden test data. Each group comprises a 1K low-light input, a GT, and a mask, maintaining a scenario distribution and degradation difficulty consistent with the Phase 2 dataset.}
    \label{fig:03}
\end{figure}

Crucially, the final inference and testing procedures were executed uniformly by the organizers in a standardized local hardware environment, directly utilizing the source codes and pre-trained model weights submitted by the top-tier participating teams. Furthermore, to prevent participants from exploiting resolution-scaling shortcuts to artificially inflate evaluation scores, we imposed a strict constraint: the spatial resolution of all model outputs must precisely match their corresponding input dimensions. Consequently, no image scaling (resizing) or spatial padding operations were permitted or introduced at any stage of the final evaluation pipeline. Two representative validation samples are illustrated in Figure \ref{fig:03}.

\subsection{Evaluation Measures}

In the context of real-world image restoration tasks, relying solely on conventional objective metrics frequently fails to adequately capture human visual perception concerning aesthetics and naturalness. To circumvent this limitation, we have engineered a comprehensive evaluation framework for this challenge, seamlessly integrating a region-aware quantitative measurement system with a rigorous expert-driven subjective evaluation protocol.

\subsubsection{Quantitative Measure}
\label{quantitativemeasure}

Considering that the AI Flash Portrait task inherently demands not only the precise restoration of intricate facial details but also the preservation of illumination consistency within the background environment, we introduced a region-aware measurement mechanism guided by the provided person masks. For both Phase 2 and Phase 3, the overall objective score is mathematically formulated as follows:

\begin{equation}
\begin{split}
    \mathit{TotalScore} &= W_{1} \cdot (1 - \mathit{LPIPS}_{person}) \\
    &\quad + W_{2} \cdot (1 - \Delta E_{person}) + W_{3} \cdot \mathit{GlobalScore} \\
    \mathit{GlobalScore} &= 0.5 \cdot \mathit{Norm}(\mathit{PSNR}_{bg}) + 0.5 \cdot \mathit{SSIM}_{global}
\end{split}
\label{eq:score}
\end{equation}

$LPIPS_{person}$ and $\Delta E_{person}$ denote the normalized perceptual similarity and color difference metrics, respectively, which are computed exclusively within the person region as delineated by the provided mask (for both metrics, lower is better). Conversely, $\mathit{PSNR}_{bg}$ and $\mathit{SSIM}_{global}$ are utilized to quantify the peak signal-to-noise ratio within the background region and the structural similarity across the entire global image, respectively (for both metrics, higher is better). The parameters $W_{1}, W_{2}, W_{3}$  serve as system-adaptive weight coefficients. Fundamentally, this composite formulation effectively prevents algorithms from artificially over-sharpening the portrait at the expense of background cleanliness, while also deterring trivial solutions that flatten facial rendering merely to inflate the global PSNR score.

\subsubsection{Subjective Evaluation}
\label{subjectivevaluation}

To guarantee the fairness and scientific rigor of the subjective evaluation, we devised a strict blind-test protocol. First, based on the objective rankings from Phase 2, we shortlisted the top 12 participating teams. Subsequently, we randomly sampled 50 groups of images from the 100 hidden test sets utilized in Phase 3. For each data group, the restored images generated by the 12 shortlisted teams were entirely anonymized and displayed in a randomized order. We assembled an independent jury comprising over five senior image processing experts and industry practitioners. The jury members were instructed to select the top-3 best-performing results from the 12 candidate images for each sample, evaluating them against the following six core dimensions:

\noindent\textbf{Facial Naturalness}. Evaluates whether the restored skin tones are healthy and accurate. Severe penalties are imposed for artificial wax-like appearances or excessive skin smoothing.

\noindent\textbf{Portrait Detail Preservation}. Assesses the authenticity and fidelity of high-frequency textural details, such as hair strands, eyelashes, and fabric patterns.

\noindent\textbf{Lighting Realism}. Examines whether the luminance distribution between the foreground and background adheres to the physical attenuation characteristics of a real flash.

\noindent\textbf{Background Cleanliness}. Judges the accuracy of background color reproduction and the effectiveness of noise suppression.

\noindent\textbf{Scene Balance}. Rewards solutions that manage to illuminate the primary subject while preserving the original ambient mood of low-light or cool-toned scenes.

\noindent\textbf{Overall Consistency}. Requires strict semantic fidelity to the input. Color banding and generative hallucinations are strictly penalized.

Upon completion of the blind test, we aggregated the total number of times each team's output was selected into the top-3 across the 50 sample groups. Finally, based on this cumulative selection frequency, the results were normalized into a subjective score ranging from 80 to 90 points. This normalized subjective score was then combined with the objective score using a 3:7 weighting ratio to determine the final, overall ranking.

\subsection{Phases}

\subsubsection{Phase 1: Model Design and Tuning}

During the model design phase, we released a comprehensive training set to the participants, comprising 600 groups of fully aligned data triplets (each containing a low-light input, the corresponding GT, and a person mask). This dataset was provided to enable participants to investigate the underlying patterns of real-world degradation and to construct their foundational models. Furthermore, the organizers released an open-source baseline model to assist competitors in rapidly establishing and validating their end-to-end development pipelines.

\subsubsection{Phase 2: Online Feedback}

In the online evaluation phase, the Phase 2 validation set was released to the participants. The competing teams were required to upload their restored images to the CodaBench evaluation platform. Upon submission, the system automatically evaluated the results utilizing the objective metrics detailed in Sec. \ref{quantitativemeasure}, subsequently updating the public leaderboard in real time. The primary objective of this phase was to provide participants with an interactive environment to validate their algorithmic hypotheses and systematically perform hyperparameter tuning.

\subsubsection{Phase 3: Final Evaluation}

Serving as the decisive phase for the final rankings, the objective online scoring system was officially closed. The top 12 teams from Phase 2 were mandated to submit their comprehensive project repositories, which included the source code, pre-trained model weights, and detailed technical documentation (specifying the training hardware configurations and the model inference time in seconds per image). To guarantee the robustness and efficiency of the winning solutions for practical industrial deployment, the organizers established the final standings through a rigorous dual-verification process: unified code reproduction conducted locally, cross-validated against the expert blind-testing protocol detailed in Sec. \ref{subjectivevaluation}.

\subsection{Awards}

The following awards of each track are provided \textbf{for each track}:\

\begin{itemize}
\item One first-class award (i.e., the champion) with a cash prize of \textbf{US\$1000};

\item Two second-class awards with cash prizes of \textbf{US\$500 each};

\item Three third-class awards with cash prizes of \textbf{US\$200 each}.
\end{itemize}

\subsection{Important Dates}

\begin{itemize}
\item 2026.01.23: Released data of phase 1. Phase 1 began;
\item 2026.01.28: Released data of phase 2. Phase 2 began;
\item 2026.03.05: Phase 3 began;
\item 2026.03.12: Phase 3 results submission deadline;
\item 2026.03.19: Final rank announced.
\end{itemize}
\section{Challenge Results}

The NTIRE 2026 AI Flash Portrait Challenge (Track 3) has garnered widespread attention from both academia and industry. Throughout the competition period, the track attracted 118 registered participating teams and accumulated 3,187 valid submissions on the CodaBench online evaluation platform. Upon entering Phase 3, we invited the top 12 teams from the Phase 2 leaderboard to submit their source code and pre-trained weights for the final comprehensive evaluation. In Section \ref{teamsandmethods1} of this paper, we detail the participating teams that advanced to the final evaluation phase along with their member affiliations, and we explore the advanced architectures and algorithmic strategies adopted by the top-tier teams.

\subsection{Phase 2: Quantitative Comparison on Validation Data}

During Phase 2, competing teams performed inference on 100 groups of validation data—where the ground truths were deliberately withheld—and submitted their predictions to the online server. The server automatically computed the scores in the background using these undisclosed GTs. The evaluation in this phase was strictly based on quantitative objective metrics, primarily relying on the region-aware measurement system proposed in Sec. \ref{quantitativemeasure}. As a reference baseline, the model provided by the organizers, trained exclusively on the Phase 1 training set, achieved a score of 82.16 on the Phase 2 online system.

The detailed objective quantitative scores of the top 12 teams in Phase 2 are presented in Table \ref{tab:01}. It is important to note that the "Phase 2 Score" represents the real-time objective score displayed on the public leaderboard during the competition. For certain teams, data fields are marked with a hyphen "-". This indicates that these teams either failed to submit valid code during the subsequent reproducibility verification stage, or the objective scores reproduced locally by the organizers critically deviated from their online leaderboard scores, thereby disqualifying their detailed sub-metric entries.

\begin{table}[t]
\caption{Phase 2 quantitative results on the validation dataset. The ``-" symbol indicates that the team either failed to submit valid code during the reproducibility check or their reproduced score deviated significantly from the online leaderboard. $\uparrow$ indicates the higher the better, $\downarrow$ indicates the lower the better.}
\label{tab:01}
\centering
\resizebox{\linewidth}{!}{
\begin{tabular}{lccccc}
\toprule
Team & $LPIPS_{person} \downarrow$ & $\Delta E_{person} \downarrow$ & GlobalScore $\uparrow$ & Phase 2 Score & Rank \\
\midrule
nunucccb & 0.0266 & 7.19 & 0.7843 & 86.10 & 2 \\
nanbei & - & - & - & 85.21 & 3 \\
SHL & 0.0268 & 6.83 & 0.7416 & 84.91 & 4 \\
kiritobryant & - & - & - & 84.90 & 5 \\
hezhaokun & 0.0270 & 6.75 & 0.7388 & 84.88 & 6 \\
KC110 & 0.0284 & 8.07 & 0.7647 & 84.33 & 7 \\
NJUST-KMG & 0.0315 & 6.88 & 0.7293 & 83.70 & 8 \\
zte\_cv & 0.0312 & 7.04 & 0.7209 & 83.41 & 9 \\
CEL-Ricky & - & - & - & 83.13 & 10 \\
lihang & - & - & - & 82.79 & 11 \\
william4s & - & - & - & 82.77 & 12 \\
\bottomrule
\end{tabular}}
\end{table}

\subsection{Phase 3:  Comprehensive Evaluation Combining Objective and Subjective Metrics}

Although the objective metrics in Phase 2 can reflect the restoration precision at the physical signal level, in the context of real-world low-light portrait computational photography, images with high objective scores do not necessarily align perfectly with advanced human aesthetic preferences. Therefore, in Phase 3, we upgraded the evaluation framework to a comprehensive system comprising "70$\%$ reproducible objective quantitative assessment + 30$\%$ expert blind-test subjective evaluation."

For the top 12 teams from Phase 2, the organizers first executed their submitted codes uniformly in a standardized local hardware environment to acquire the authentic objective scores of each model on the Phase 3 hidden test set. Subsequently, we rigorously implemented the subjective evaluation protocol detailed in Sec. \ref{subjectivevaluation}: We randomly sampled 50 groups of predictions from the Phase 3 test set, entirely anonymizing and randomly shuffling the images generated by the 12 teams on a single screen. An independent jury composed of senior image algorithm experts evaluated the images based on six core dimensions—such as "Facial Naturalness," "Lighting Realism," and "Background Cleanliness"—and selected the top-3 best-performing images for each sample group. The system aggregated the total number of times each team's output was selected into the top-3 and linearly normalized these cumulative frequencies into a subjective score (UScore) ranging from 80 to 90 points.

The Final Score was calculated as the weighted sum of the newly reproduced objective score (70$\%$ weight) and the subjective UScore (30$\%$ weight). Following this rigorous dual-verification process, the final scores and rankings of the top 12 winning teams are summarized in Table \ref{tab:phase3_final_results}.

\begin{figure*}[t]
    \centering
    
    
    
    \includegraphics[width=\textwidth]{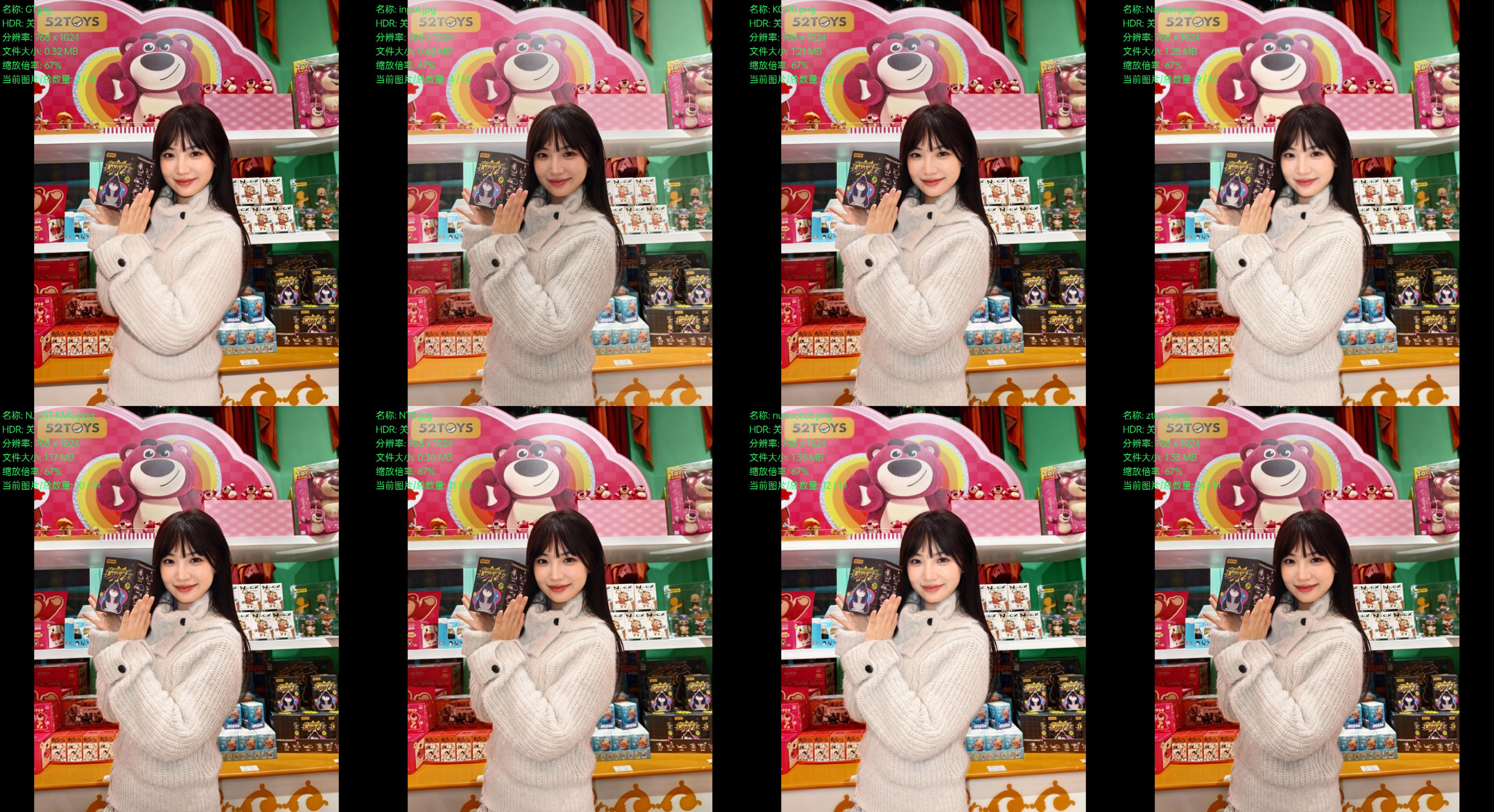}
    
    \caption{Visual comparison of the restoration results from the top 6 teams. The restored images are evaluated against the low-light input and the high-quality GT in terms of facial naturalness, detail preservation, lighting realism, and background cleanliness.}
    \label{fig:result}
\end{figure*}

From the final comprehensive evaluation results of Phase 3, a critical phenomenon can be observed: several models that excelled in the single objective metrics during Phase 2 experienced significant fluctuations in their rankings after the introduction of the expert subjective blind test. This corroborates our primary motivation for organizing this challenge—a non-negligible domain gap still exists between traditional image quality assessment metrics and human perception in complex, real-world portrait generation tasks. The ultimate winning solutions not only remained faithful to the GTs in terms of feature space and pixel distribution but also demonstrated exceptional industrial deployment potential across "soft aesthetic" dimensions, such as healthy skin tone rendering, logical physical light attenuation, and immersive ambient scene mood preservation. The visual effects of the top six ranked images are shown in Figure \ref{fig:result}.

\begin{table}[t]
\caption{Final comprehensive results of Phase 3. The Final Score is computed as the weighted sum of the reproduced Objective Score (70\%) and the expert blind-test UScore (30\%). $\uparrow$ indicates the higher the better, $\downarrow$ indicates the lower the better.}
\label{tab:phase3_final_results}
\centering
\resizebox{\linewidth}{!}{
\begin{tabular}{lcccc}
\toprule
Team & Objective Score & UScore & Final Score & Rank  \\
\midrule
nunucccb & 85.81 & 89.09 & 86.794 & 1 \\
nanbei & 84.66 & 90.00 & 86.262 & 2 \\
KC110 & 85.37 & 87.27 & 85.940 & 3 \\
NJUST-KMG & 83.91 & 88.18 & 85.191 & 4 \\
zte\_cv & 84.28 & 84.55 & 84.361 & 5 \\
NTR & 82.94 & 86.36 & 83.966 & 6 \\
hezhaokun & 83.90 & 83.64 & 83.822 & 7 \\
CEL-Ricky & 82.61 & 85.45 & 83.462 & 8 \\
lihang & 82.97 & 83.64 & 83.171 & 9 \\
SHL & 83.47 & 81.82 & 82.975 & 10 \\
kiritobryant & 81.78 & 80.00 & 81.246 & 12 \\
\bottomrule
\end{tabular}}
\end{table}
\section{Teams and Methods}
\label{teamsandmethods1}
Due to space limitations, we describe their proposed methods for this track in the supplementary material.
\section{Acknowledgments}

We extend our sincere gratitude to all the human subjects who participated in the data collection process. The data acquisition protocol rigorously adhered to established ethical standards, and explicit Model Release Agreements were obtained from all featured individuals to ensure full legal compliance and the protection of their portrait rights. Importantly, we must emphatically state that this dataset is strictly restricted to non-commercial, academic research purposes only. Any form of commercial exploitation, malicious manipulation (e.g., DeepFake generation), or unauthorized redistribution is strictly prohibited.

We also thank the NTIRE 2026 sponsors, including OPPO and the University of W\"urzburg, for their support.

\section{Appendix: Teams and affiliations}
\label{appendix}

\textbf{NTIRE 2026 Team}

\noindent\textit{\textbf{Challenge:}} 

\noindent NTIRE 2026 The 3rd Restore Any Image Model (RAIM): AI Flash Portrait (Track 3)

\noindent\textit{\textbf{Organizers:}}

\noindent Ya-nan Guan$^{1,2,3}$ (guanyanan@mail.nankai.edu.cn) 

\noindent Shaonan Zhang$^{1,4}$ (13414561874@163.com)

\noindent Hang Guo$^{3}$ (cshguo@gmail.com) 

\noindent Yawen Wang$^{4}$ (19137539336@163.com)

\noindent Xinying Fan$^{4}$ (13178173390@163.com)

\noindent Tianqu Zhuang$^{3}$ (zhuangtq23@mails.tsinghua.edu.cn)

\noindent Jie Liang$^{1}$ (liang27jie@163.com)

\noindent Hui Zeng$^{1}$ (cshzeng@gmail.com)

\noindent Guanyi Qin$^{1, 6}$ (guanyi.qin@u.nus.edu)

\noindent Lishen Qu$^{1, 2}$ (qulishen@mail.nankai.edu.cn)

\noindent Prof. Tao Dai$^{4}$ (daitao@szu.edu.cn)

\noindent Prof. Shu-Tao Xia$^{3}$ (xiast@sz.tsinghua.edu.cn)

\noindent Prof. Lei Zhang$^{1,5}$ (cslzhang@comp.polyu.edu.hk)

\noindent Prof. Radu Timofte$^{7}$ (radu.timofte@uni-wuerzburg.de)

\noindent\textit{\textbf{Affiliations:}}

\noindent $^1$ OPPO Research Institute, China

\noindent $^2$ Nankai University, China

\noindent $^3$ Tsinghua University, China

\noindent $^4$ Shenzhen University, China

\noindent $^5$ The Hong Kong Polytechnic University, China

\noindent $^6$ National University of Singapore, Singapore

\noindent $^7$ Computer Vision Lab, University of W\"urzburg, Germany

~\\

\noindent\textit{\textbf{Team name:}}

\noindent\textit{\textbf{Team name:}} nunucccb

\noindent\textit{\textbf{Members:}}  
Bin Chen (nunucccb@gmail.com), Yuanbo Zhou, Hongwei Wang, Qinquan Gao, Tong Tong

\noindent\textit{\textbf{Affiliations:}}  
Fuzhou University, Imperial Vision Technology

~\\

\noindent\textit{\textbf{Team name:}} nanbei

\noindent\textit{\textbf{Members:}}  
Yanxin Qian (984784576@qq.com), Lizhao You (lizhaoyou@xmu.edu.cn)

\noindent\textit{\textbf{Affiliations:}}  
School of Informatics, Xiamen University

~\\

\noindent\textit{\textbf{Team name:}} KC110

\noindent\textit{\textbf{Members:}}  
Jingru Cong (congjingru0412@gmail.com), Lei Xiong, Shuyuan Zhu

\noindent\textit{\textbf{Affiliations:}}  
University of Electronic Science and Technology of China

~\\

\noindent\textit{\textbf{Team name:}} NJUST-KMG

\noindent\textit{\textbf{Members:}}  
Zhi-Qiang Zhong (1533534827@qq.com), Kan Lv, Yang Yang

\noindent\textit{\textbf{Affiliations:}}  
Nanjing University of Science and Technology

~\\

\noindent\textit{\textbf{Team name:}} zte\_cv

\noindent\textit{\textbf{Members:}}  
Kailing Tang (tang.kailing@zte.com.cn), Minjian Zhang, Zhipei Lei, Zhe Xu, Liwen Zhang, Dingyong Gou, Yanlin Wu, Cong Li, Xiaohui Cui

Minjian Zhang, Liwen Zhang, Zhe Xu, Zhipei Lei

\noindent\textit{\textbf{Affiliations:}}  
Zhongxing Telecom Equipment

\noindent\textit{\textbf{Codabench Username:}} sky\_flight

~\\

\noindent\textit{\textbf{Team name:}} NTR

\noindent\textit{\textbf{Members:}}  
Jiajia Liu, Guoyi Xu, Yaoxin Jiang, Yaokun Shi, Jiachen Tu (jtu9@illinois.edu)

\noindent\textit{\textbf{Affiliations:}}  
University of Illinois Urbana-Champaign

\noindent\textit{\textbf{Codabench Username:}} miketjc

~\\





\noindent\textit{\textbf{Team name:}} CEL-Ricky

\noindent\textit{\textbf{Members:}}  
Liqing Wang (22230002@zju.edu.cn)

\noindent\textit{\textbf{Affiliations:}}  
Color and Engineering Lab, Zhejiang University

~\\





\noindent\textit{\textbf{Team name:}} SHL

\noindent\textit{\textbf{Members:}}  
Shihang Li (lishihang@gml.ac.cn), Bo Zhang, Biao Wang

\noindent\textit{\textbf{Affiliations:}}  
Guangdong Laboratory of Artificial Intelligence and Digital Economy (SZ), Shenzhen, China

~\\






\noindent\textit{\textbf{Team name:}} kiritobryant

\noindent\textit{\textbf{Members:}}  
Haiming Xu (24011211044@stu.xidian.edu.cn), Xiang Long, Xurui Liao, Yanqiao Zhai, Haozhe Li

\noindent\textit{\textbf{Advisor:}} Qianqian Wang

\noindent\textit{\textbf{Affiliations:}}  
Xidian University

~\\

\noindent\textit{\textbf{Team name:}} APRIL-AIGC

\noindent\textit{\textbf{Members:}}  
Shijun Shi, Jiangning Zhang, Yong Liu, Kai Hu, Jing Xu, Xianfang Zeng

\noindent\textit{\textbf{Affiliations:}}  
Jiangnan University; Zhejiang University; University of Science and Technology of China

~\\

\noindent\textit{\textbf{Team name:}} Cody98

\noindent\textit{\textbf{Members:}}  
Yuyang Liu, Minchen Wei (minchen.wei@polyu.edu.hk)

\noindent\textit{\textbf{Affiliations:}}  
The Hong Kong Polytechnic University
{
    \small
    \bibliographystyle{ieeenat_fullname}
    \bibliography{main}
}


\end{document}